\documentclass[a4paper,twoside]{article}

\usepackage{epsfig}
\usepackage{calc}
\usepackage{amssymb}
\usepackage{amstext}
\usepackage{amsmath}
\usepackage{amsthm}
\usepackage{multicol}
\usepackage{pslatex}
\usepackage{hyperref} 
\usepackage{ICAIIT}   

\usepackage{xcolor}
\usepackage{color}
\usepackage{tabularray}
\usepackage{subfig}
\usepackage{rotating}
\usepackage{cite}
\usepackage{amsmath,amssymb,amsfonts}
\usepackage{algorithmic}

\usepackage{textcomp}
\usepackage{xcolor}

\usepackage{subcaption}

\usepackage{amsmath}
\usepackage{amssymb}

\usepackage{graphicx}
\usepackage{amsmath}
\usepackage{multirow}
\usepackage{changepage}
\usepackage{footnote}
\usepackage[marginal]{footmisc}

\begin{document}

\title{Semi-Supervised Learning for Anomaly Traffic Detection via Bidirectional Normalizing Flows}

\author{\authorname{Zhangxuan~Dang\sup{1}, Yu~Zheng\sup{1},Xinglin~Lin\sup{1},Chunlei~Peng\sup{1},
Qiuyu Chen\sup{2},
Xinbo~Gao\sup{3}}
\affiliation{\sup{1} Xidian University}
\affiliation{\sup{2}Amazon}
\affiliation{\sup{3}Chongqing University of Posts and Telecommunications}
}

\abstract{With the rapid development of the Internet, various types of anomaly traffic are threatening network security. We consider the problem of anomaly network traffic detection and propose a three-stage anomaly detection framework using only normal traffic. Our framework can generate pseudo anomaly samples without prior knowledge of anomalies to achieve the detection of anomaly data. Firstly, we employ a reconstruction method to learn the deep representation of normal samples. Secondly, these representations are normalized to a standard normal distribution using a bidirectional flow module. To simulate anomaly samples, we add noises to the normalized representations which are then passed through the generation direction of the bidirectional flow module. Finally, a simple classifier is trained to differentiate the normal samples and pseudo anomaly samples in the latent space. During inference, our framework requires only two modules to detect anomalous samples, leading to a considerable reduction in model size. According to the experiments, our method achieves the state of-the-art results on the common benchmarking datasets of anomaly network traffic detection. The code is given in the https://github.com/ZxuanDang/ATD-via-Flows.git}

\onecolumn \maketitle \normalsize \vfill

\section{Introduction}
With the development of the Internet, the proliferation of devices has led to explosive growth in the Internet traffic, which poses significant challenges to the management of network resources and the assurance of network security.
In particular, the increasing complexity and diversity of network attacks require systems to enhance their ability to detect anomaly traffic.
Anomaly network traffic detection is a vital component in ensuring network security by detecting anomaly traffic passing through computer network nodes. Such network traffic may include malicious activity that is not in alignment with normal behavior.
It is critical to maintaining the security of the network infrastructure and reduces the likelihood of network intrusions.

Supervised methods are used to detect anomaly traffic~\cite{salman2022machine,niu2019abnormal,gao2020malicious,li2017intrusion,yang2020griffin,zheng2022multi}. 
For example, a machine learning classification model, trained on appropriately labelled manual features, will declare anomaly traffic when the data does not follow the normal distribution.
However, the main drawbacks of supervised anomaly detection are~\cite{chandola2009anomaly,jadidi2015flow,bhuyan2014towards,shi2022unsupervised}:
(1) Collecting anomaly traffic would be a time-consuming and labor-intensive task due to the nature of the anomaly traffic;
(2) It can be challenging to obtain accurate and representative labels for normal and abnormal traffic.
Due to limited access to a large amount of anomaly data, semi-supervised methods are often adopted for detecting anomalies by training only on normal traffic.

\begin{figure}[t]
\begin{center}
\includegraphics[width=0.95\linewidth]{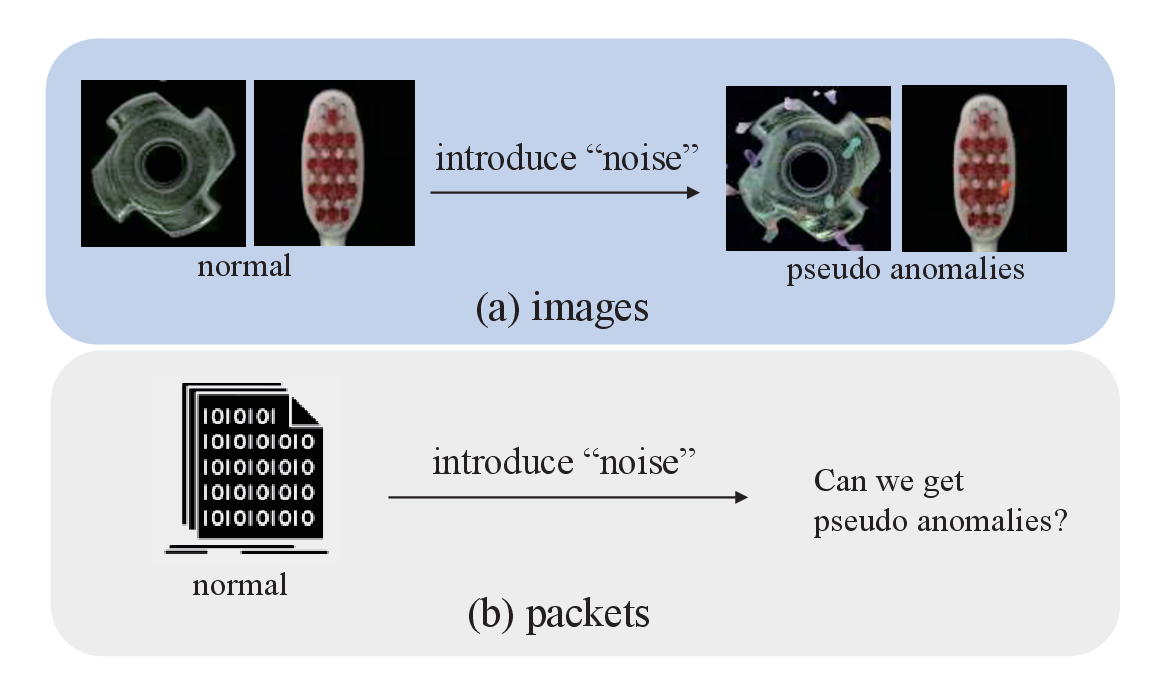}
\end{center}
   \caption{(a) Anomalies in images comprise of both colour and shape. Based on prior knowledge of anomaly patterns, images can simulate anomalies by introducing "noise"~\cite{li2021cutpaste,zavrtanik2021draem}.
   (b)
   Network traffic anomaly patterns are difficult to generalise. Simulating abnormal network traffic packets by directly introducing "noise" may destroy the semantic information of the data packets and produce meaningless pseudo anomalies, as shown in Section~\ref{ablation study}. Our framework is able to simulate anomaly samples without prior knowledge of anomaly patterns.
   }
   \label{fig:motavation}
\end{figure}

Alternative methods generate network traffic to address data labeling and scaling issues.
For example, Ring et al. propose three different preprocessing methods for GANs that generate flow-based data and evaluate the quality of the generated flows~\cite{ring2019flow}. 
In data imputation, SS-GACN and GACN allow for missing values in data labels and features~\cite{ghanavi2021generative}.
The methods impute the missing data features based on classification accuracy.
Both of these methods demonstrate that GANs can successfully generate real network traffic.
However, the generation of network traffic by GANs necessitates large-scale data. Moreover, such techniques can only generate network traffic with a distribution similar to the collected data, making it challenging to simulate anomaly traffic from diverse distributions~\cite{cheng2019pac}.  

In this work, we propose a novel method for simulating anomalies that uses only normal traffic during training. Our method can generate anomaly samples of network traffic without any prior knowledge of the anomalies, thereby improving anomaly detection.
Anomaly simulation-based methods are often used for anomaly detection in computer vision~\cite{bergman2020classification,golan2018deep,hendrycks2019using}.
As shown in Figure~\ref{fig:motavation}, they generate new data outside the distribution of normal data by applying transformations such as rotation, CutPaste~\cite{li2021cutpaste}, flipping, and Cutout~\cite{devries2017improved} to the original normal images, and then classify the data using a classifier.
It has been proven that this approach can successfully distinguish between normal and anomaly samples~\cite{li2021cutpaste}.

Using the prior knowledge of anomaly patterns, geometric transformation enables the generation of anomaly samples by altering the colour and shape of normal images.
For network traffic packets, it is difficult to simulate anomaly patterns with transformations.
We cannot directly use geometric transformations such as Cutout to obtain anomaly samples, as the packets are one-dimensional data structures with precise semantics and no spatial semantics, producing meaningless pseudo anomalies, as shown in Section~\ref{ablation study}.
Therefore, a bidirectional flow module is proposed in our method.
This module can normalize the normal packet feature to a specific tractable distribution.
The unknown anomaly samples will be outside this distribution after normalization, as the experiments show.
By manipulating the vectors in the distribution, we are able to change the properties of the samples, enabling the simulation of anomalies~\cite{kingma2018glow}. 
By introducing noise to the normalized features of normal samples, we can make them deviate from the distribution of normal samples. 
Then, through the direction of the flow generation, we can map them back to the original space to generate anomaly samples.
Our framework introduces random noise to achieve simulation of anomaly samples, provided the anomaly pattern is unknown.
By conducting a proxy classification task between the normal samples and the synthetic ones, we facilitate the model in accurately identifying normal samples.
As shown in Section~\ref{others}, pseudo anomaly samples help the model to better detect normal representations, even if they are almost not overlapped with real anomaly traffic.
To the best of our knowledge, this is the first time that a normalizing flow module has been used to generate anomaly traffic network samples with no prior knowledge of anomaly patterns and has led to good results in anomaly detection.

In summary, the main contributions of this paper are in three folds:

\begin{itemize}
    \item 
    This paper introduces the normalizing flows to formulate a three-stage framework for anomaly traffic detection by using only normal traffic data.
    The normalizing flows are utilized to process the packets obtained from the feature extractor.
    The exceptional performance observed in downstream anomaly detection demonstrates the potential usage for adopting normalizing flows in anomaly traffic detection.
    \item 
    This paper embeds the normalizing flows into the process of generating anomaly traffic samples with no prior knowledge of anomaly patterns.
    The proposed bidirectional flow module effectively utilizes both normalization and generation directions to simulate anomaly samples by manipulating the normalized vector without prior knowledge of the anomaly patterns.
    The detection results demonstrate a significant improvement in performance achieved through the simulated anomaly samples.
    \item 
    Our method outperforms other popular anomaly detection methods on three common benchmarking datasets for anomaly network traffic detection and is efficient in computation.
\end{itemize}

\section{Related Work}
\label{gen_inst}

\paragraph{Anomaly Network Traffic Detection Methods}
In the current research on anomaly network traffic detection, deep learning-based methods are widely used.
Some researchers combine traditional feature extraction with the classification ability of neural network models.
Cao et al.~\cite{cao2022network} utilize the RFP algorithm for traffic feature extraction and incorporate convolutional neural networks (CNNs) and Gated Recurrent Units (GRU) for classification.
Saba et al.~\cite{saba2022anomaly} utilize a CNN-based classification approach to predict anomaly traffic based on the features of traffic datasets.
Liu et al.~\cite{liu2019ddos} utilize a BP neural network model to detect manually extracted flow features.

Other researchers leverage the powerful feature extraction capability of neural network models to extract features from traffic data, thereby enhancing the performance of classifiers.
Shone et al.~\cite{shone2018deep} utilize a non-symmetric autoencoder for feature extraction and subsequently integrate it with random forests for intrusion detection.
Javaid et al.~\cite{javaid2016deep} propose a sparse autoencoder to learn feature extraction from unlabeled data, effectively leveraging the available data to enhance the feature extraction capability. Subsequently, they apply a classification task for detection purposes.
Wang et al.~\cite{wang2017hast} employ CNNs and long short-term memory (LSTM) networks to effectively learn spatial and temporal features for classification.
These methods are all fully supervised learning methods, which require collecting a large amount of labeled anomaly traffic. In contrast, our method achieves effective anomaly detection by leveraging easily collectible normal data without the need for labeled anomaly traffic.

\paragraph{Anomaly Detection Methods}
The research on anomaly detection encompasses various methods, including reconstruction-based, feature matching-based, and anomaly simulation-based approaches.
Reconstruction-based methods aim to detect anomaly samples through the analysis of reconstruction errors.
Akcay et al.~\cite{akcay2019ganomaly} detect anomaly images by reconstructing the latent vectors using encoders.
Feature matching-based methods calculate the difference between test samples and stored embeddings to detect anomaly samples.
Roth et al.~\cite{roth2022towards} obtain anomaly scores by calculating the distance between test samples and normal embeddings stored in a memory bank.
Simulation anomaly sample-based methods utilize the synthetic anomaly samples to enhance the feature extraction or make the models clearly distinguish the differences between normal and anomaly samples.
Some researches use geometric transformations to simulate anomaly samples\cite{golan2018deep,hendrycks2019using,li2021cutpaste,song2021anoseg}, other researches simulate anomaly samples by adding noise~\cite{zavrtanik2021draem,yang2023memseg,collin2021improved}.
The experimental results in these papers show that simulated anomaly samples enhance the overall detection performance.
However, these methods directly process images to obtain synthetic anomaly samples, which can not be applied to network traffic.
Our method can simulate network traffic anomaly samples and demonstrates excellent performance in anomaly traffic detection.

\paragraph{Traffic Generation Methods}
In the current work on network traffic analysis, there are many studies that use traffic generation to improve the performance of detection systems.

On the one hand, in research on preventing adversarial sample attacks, traffic generation techniques are used to generate adversarial samples to improve the robustness and accuracy of the model.
Elie et al.~\cite{alhajjar2021adversarial} focus on the attack perspective and investigate techniques to generate adversarial examples that can evade machine learning models. They specifically explore the use of evolutionary computation and deep learning as tools for adversarial example generation.
Ye et al.~\cite{peng2020detecting} propose a defense algorithm using a bidirectional generative adversarial network (GAN) to improve the robustness and accuracy of NIDS in the adversarial environment. The algorithm involves training the generator to learn the data distribution of normal samples and using the discriminator to detect adversarial samples based on reconstruction and matching errors.
Wang et al.~\cite{wang2021def} proposed Def-IDS mechanism is a two-module training framework that integrates multi-class generative adversarial networks and multi-source adversarial retraining to improve model robustness while maintaining detection accuracy on unperturbed samples.
Zolbayar et al.~\cite{zolbayar2022generating} develop a generative adversarial network (GAN)-based attack algorithm called NIDSGAN to generate realistic adversarial network traffic flows that can evade ML-based NIDS.
The main contributions of the paper~\cite{abdelaty2021gadot} are the proposal of GADoT, an adversarial training framework that leverages GANs to generate fake-benign samples for perturbing DDoS samples, and the evaluation of GADoT using network traffic traces capturing adversarially perturbed SYN and HTTP DDoS flood attacks.

On the other hand, in the work of network traffic classification, traffic generation techniques are used to perform data augmentation on the traffic data to synthesise network packets that are as realistic as possible.
Shahid et al.~\cite{shahid2020generative} propose combining an autoencoder with a Generative Adversarial Network (GAN) to generate sequences of packet sizes that mimic the behavior of real bidirectional flows. The autoencoder is trained to learn a latent representation of real packet size sequences, and the GAN is trained on the latent space to generate realistic sequences.
Nukavarapu et al.~\cite{nukavarapu2022miragenet} introduce MirageNet, a GAN-based framework for synthetic network traffic generation. The first component of MirageNet, MiragePkt, validates the performance of their framework using synthesized DNS packets.
Yin et al.~\cite{yin2022practical} propose an end-to-end framework called NetShare to explore the feasibility of using Generative Adversarial Networks (GANs) to generate synthetic packet and flow header traces for networking tasks.
Hui et al.~\cite{hui2022knowledge} propose a knowledge-enhanced generative adversarial network (GAN) framework to generate realistic IoT traffic. The framework incorporates semantic knowledge and network structure knowledge of various IoT devices through a knowledge graph.

Generative Adversarial Networks (GANs) are frequently utilised in all of these approaches to generate network traffic. However, generating network traffic through GANs demands a vast amount of training samples, and the procurement of malicious traffic can be challenging.
In addition, GANs can only generate samples that are similar to the training set, making it difficult to generate data outside of the training distribution~\cite{cheng2019pac}.

\begin{figure*}[htb]
\begin{center}
\includegraphics[width=1\linewidth]{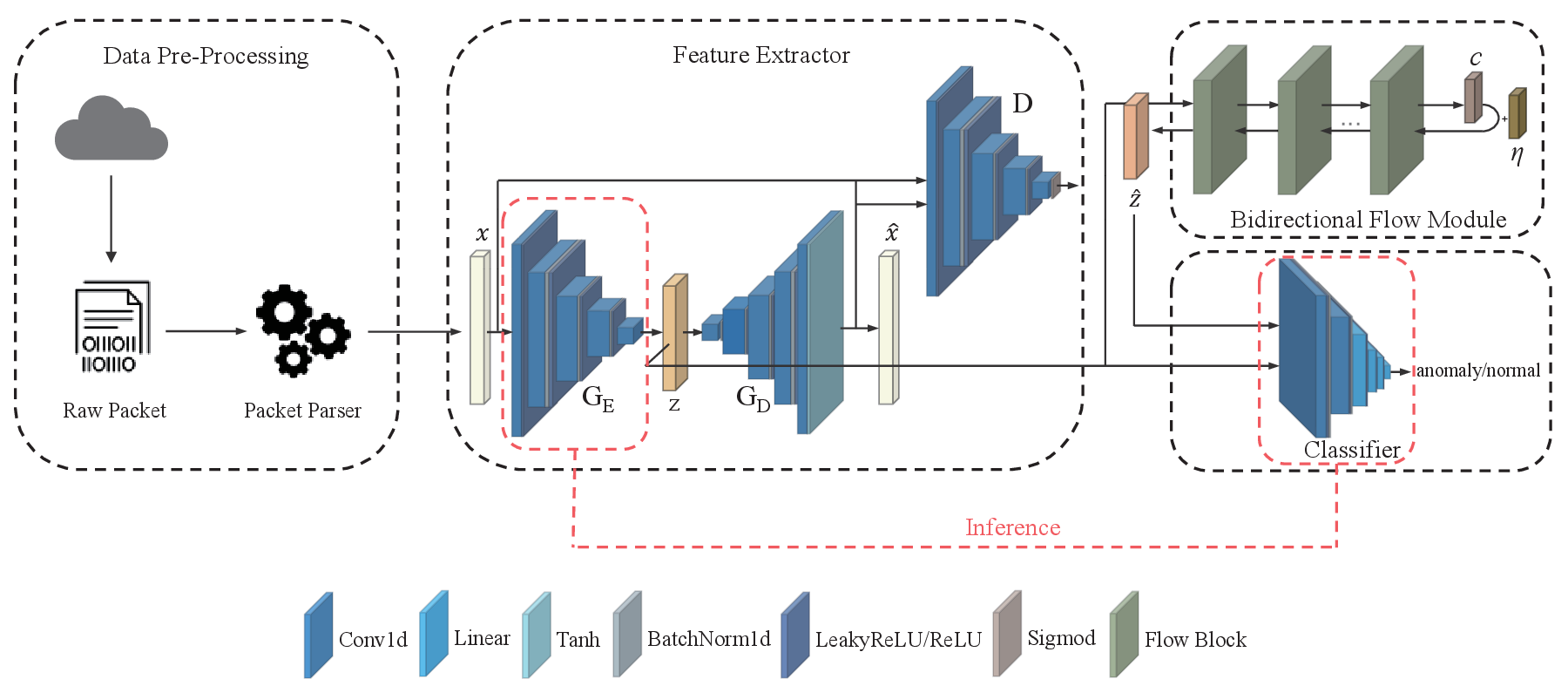}
\end{center}
    \caption{An overview of our framework for anomaly detection. $c$ corresponds to the representation of normal packets in the standard normal space, and $\eta$ corresponds to the noise vector sampled from a Gaussian Distribution. Feature Extractor is trained to perform deep feature extraction on one-dimensional normal packets. Bidirectional Flow Module is trained to normalize the representation of normal packets to a standard normal distribution. During the training of Classifier, the representation of normal packets is normalized to a standard normal distribution. In the standard normal space, we introduce noise sampled from a Gaussian Distribution to the normalized representation, and then simulate the representation of anomaly traffic through the generation direction. Classifier is trained to distinguish the representation of normal packets $z$ and the simulated representation of anomaly packets $\hat{z}$, enabling efficient anomaly detection. In the inference phase, our method can achieve good anomaly detection by maintaining only two modules, which greatly reduces the size of the model.
    }
    \label{fig:model}
\end{figure*}

\section{Proposed Method}
\label{headings}
As shown in Figure~\ref{fig:model}, our framework is developed in three stages: feature extractor, bidirectional flow module, and classifier. In the following, we will introduce each stage.

\subsection{Feature Extractor}
The traffic packets is a one-dimensional structure with different protocol layers. By following the pre-processing steps showed in Section~\ref{section:pre-processing}, the headers of Network Layer and Transport Layer as well as the payloads are the one-dimensional vector input for the model.
Intuitively speaking, an extracted feature vector that effectively represents the traffic packet is crucial for the development of downstream tasks, because the original packet often contains a significant amount of redundant information that can confuse the model.
Related experiments will be described in Section~\ref{ablation study}.
In the field of computer vision, pre-trained models are often used when feature extraction is required, such as ResNet18 and ResNet50.
To perform feature extraction on original traffic packets, we pre-trained the feature extractor using our datasets of normal samples. 

As we only have normal samples, we use a reconstruction method to extract features.
Our feature extractor is composed of a generator $G$ and a discriminator $D$, as illustrated in Figure \ref{fig:model}.
The generator $G$ is further divided into an encoder $G_{E}$ and a decoder $G_{D}$.

Similar to \cite{akcay2019ganomaly}, we adopt both reconstruction loss and adversarial loss for better reconstruction training.
The training objectives of the model are as follows:

\small
\begin{equation}
\begin{aligned}
       \mathcal{L}_{G} = \ & \omega _{a d v} \mathbb{E}_{x \sim p_{\mathbf{X}}}\|D(x)-D(G(x))\|_{2}^{2} \\&+\omega _{r e c} \mathbb{E}_{x \sim p_{\mathbf{X}}}\|x-G(x)\|_{2}^{2}   
\end{aligned}
\end{equation}

\small
\begin{equation}
\begin{aligned}
    \mathcal{L}_{D}= \mathbb{E}_{x \sim p_{\mathbf{X}}} [ 1-D(x) ] + \mathbb{E}_{x \sim p_{\mathbf{X}}} [ D(G(x)) ]
\end{aligned}
\end{equation}

where $\hat{x}=G(x)$.

We use one-dimensional traffic packets $x,x\in  R^{n} $ as the input for our model. 
The encoder $G_{E}$ compresses the input $x$ into a hidden vector $z$, while the decoder $G_{D}$ outputs $\hat{x}$, which is responsible for reconstructing the hidden vector $z$ back to the input packets.
We believe that if the reconstructed $\hat{x}$ by the model is very close to the input $x$, then the hidden vector $z$ can effectively represent the features of the network packet.
After pre-training, we retain only the $G_{E}$ of the feature extractor for extracting features from the traffic packets, resulting in a significant reduction in the model size.

\subsection{Bidirectional Flow Module}
The normalizing flows contain both normalization and generation directions.
Generative flow models leverage a sequence of invertible and differentiable operations to transform a simple and tractable distribution into a complex distribution~\cite{dinh2016density}.
Generally, it can be described by the following formula:

\begin{equation}
    					c\sim P_{\theta }\left ( c \right )
\end{equation}
\begin{equation}
                            z=g_{\theta } \left ( c \right )
\end{equation}

where $c$ is a random vector that follows a distribution $P_{\theta }$ , typically $P_{\theta }$ is a simple distribution such as a standard normal distribution.
$z$ is a extracted vector which follows the unknow true data distribution $P^{*}(z)$.
The function $g_{\theta }$ is a reversible and differentiable function that can generate real samples from the complex distribution by utilizing samples from a simple distribution.
This direction is often called the generation direction;
The function $f_{\theta }$ is the inverse function of $g_{\theta }$, $f_{\theta }=g_{\theta } ^{-1} $, which normalizes the real samples from the complex distribution into the space of a simple distribution.
This direction is often called the normalization direction.

According to \emph{the change of variables}, the probability density of the vector $z$ can be written in the following form:

\begin{equation}
\label{max}
\log p_{{\theta}}(\mathbf{z})=\log p_{{\theta}}(\mathbf{c})+\log|\det(d\mathbf{c}/d\mathbf{z})|
\end{equation}

where $\log|\det(d\mathbf{c}/d\mathbf{z})|$ refers to the logarithm of the absolute value of the determinant of the Jacobian matrix ($d\mathbf{c}/d\mathbf{z}$), which can be easily calculated using matrix transformations~\cite{kingma2016improved}, such as triangular matrix transformations.

In fact, it is difficult to construct a powerful, reversible, differentiable and easy-to-calculate Jacobian function~\cite{kobyzev2020normalizing}. 
So in the normalizing flows,  it is common to combine a sequence of reversible and differentiable functions to achieve the desired transformation.
This is the reason why this approach is referred to as a "flow".
Therefore, the transform between $c$ and $z$ can be written as:

\begin{equation}
    \mathbf{z} \stackrel{\mathbf{g}_{1}}{\leftarrow} \mathbf{h}_{1} \stackrel{\mathbf{g}_{2}}{\leftarrow} \mathbf{h}_{2} \cdots \stackrel{\mathbf{g}_{K}}{\leftarrow} \mathbf{c}
\end{equation}
\begin{equation}
\mathbf{z} \stackrel{\mathbf{f}_{1}}{\rightarrow} \mathbf{h}_{1} \stackrel{\mathbf{f}_{2}}{\rightarrow} \mathbf{h}_{2} \cdots \stackrel{\mathbf{f}_{K}}{\rightarrow} \mathbf{c}
\end{equation}

We follow the work in~\cite{dinh2016density} and employ the affine coupling layers in each block of the bidirectional flow module, as showed in Figure~\ref{fig:model}.
Our bidirectional flow module takes the feature extracted by the feature extractor as input,
and we maximize Eq.~\ref{max} to train this module through the normalization direction.
After training, the bidirectional flow module can map the features of normal samples to the standard normal distribution space $V$.
For normal samples, the flow module maps their features to the standard normal distribution. However, the situation is different for anomaly samples. During the feature extraction phase, the pre-trained feature extractor has only been trained on normal samples. As a result, the features extracted from anomaly samples are likely to deviate from the distribution of normal features.
The bidirectional flow module is also trained on normal samples, which means that the anomaly features, deviating from the normal sample distribution may fall outside the standard normal distribution in the space $V$, as showned in Figure~\ref{fig:t-sne}.
To simulate the distribution of anomaly samples, we introduce noise into normal samples in standard normal space $V$.
For the sake of simplicity, we choose to randomly sample from Gaussian Distributions to generate noise.
We employ a reparameterization trick to represent noise:

\begin{equation}
    \eta =\mu + \sigma \odot \varepsilon
\end{equation}

where $\mu$ represents the mean vector, $\sigma$ represents the standard deviation vector, and $\varepsilon$ is the random noise sampled from the standard normal distribution.
By adjusting $\mu$ and $\sigma$, we can generate noise samples from different Gaussian Distributions.

Then we use the generation direction of the bidirectional flow module to transform the simulated anomaly samples in the normal distribution space $V$, resulting in vectors in the latent space.
In this way, we have the representation vector of normal samples and anomaly samples in the latent space.
In the field of computer vision, operations such as Cutpaste~\cite{li2021cutpaste} are applied to images of normal samples to obtain synthetic anomaly samples.
However, when it comes to traffic packets, it is challenging to obtain anomaly samples that closely resemble the real network environment through image processing techniques. This is because traffic packets lack spatial semantic information.
Therefore, it is necessary to use the bidirectional flow module to map the feature vectors to the standard normal space $V$ to construct anomaly samples. 
In the standard normal space $V$, we can manipulate the attributes of the vectors to bring them closer to real anomaly traffic samples in the latent space. This capability allows for the generation of synthetic anomaly samples that exhibit similarity to real-world anomaly traffic patterns.

\subsection{Classifier}

After obtaining normal and pseudo anomaly sample features, it is natural and straightforward to consider using a classifier for anomaly detection.
We only employ a simple classifier to classify the obtained normal and anomaly feature vectors.
We aim to improve the detection performance of real anomaly samples by encouraging the classifier to focus more on real normal samples.
To achieve this, we reduce the number of anomaly samples to half that of the normal samples.
This prevents the model from overemphasising the features of the pseudo anomaly samples.
In the testing phase, we only need to combine the classifier and encoder $G_{E}$ for anomaly detection, which significantly reducing the model's parameters and enhancing the model's deployment flexibility.
We detect anomaly traffic based on the classification results.

\begin{figure*}[htb]
  \centering
  \subfloat[Tor and non-Tor dataset]{
    \includegraphics[width=0.3\textwidth]{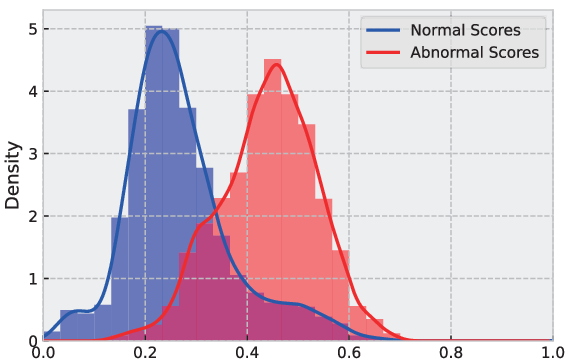}
  }
  \subfloat[VPN and non-VPN dataset]{
    \includegraphics[width=0.3\textwidth]{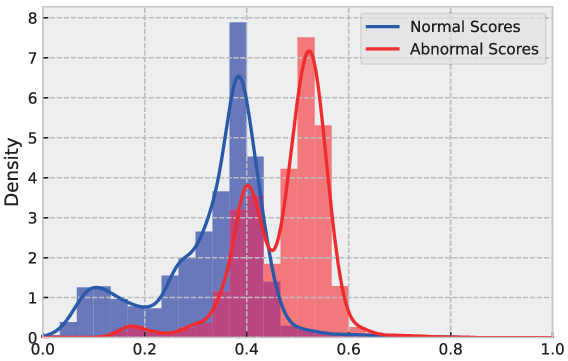}
  }
  \subfloat[DataCon2020 dataset]{
    \includegraphics[width=0.3\textwidth]{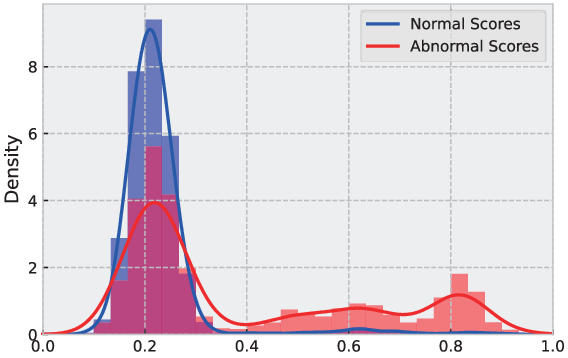}
  }
  \caption{Histogram with density curve. We plot the detection result histogram of the samples in the testing sets of the three datasets. The curve represents the kernel density estimation of the results. Our method is more effective in distinguishing between anomaly and normal traffic 
  on the "UNB-CIC Tor and non-Tor" and the "ISCX VPN and non-VPN" datasets. Although the results of distinguishing normal and anomaly traffic on the "DataCon2020" dataset are not satisfactory, they are still better than those achieved by other methods.
  }
  \label{fig:anomaly_score}
\end{figure*}

\begin{table}
\scalebox{0.9}{
\centering
\begin{tblr}{
  vline{2} = {-}{},
  hline{1-2,13-14} = {-}{},
}
Methods                & VPN    & TOR    & DataCon \\
Reverse Distillation~\cite{deng2022anomaly} & 0.6116 & 0.7450  & 0.6762  \\
DFKDE~\cite{anomalib}       & 0.5907 & 0.7356 & 0.3969  \\
DFM~\cite{ahuja2019probabilistic}                   & 0.7156 & 0.7514 & 0.6744  \\
DRAEM~\cite{zavrtanik2021draem}                 & 0.5698 & 0.7028 & 0.6479  \\
FastFlow~\cite{yu2021fastflow}              & 0.6195 & 0.6689 & 0.6571  \\
PADIM~\cite{defard2021padim}                 & 0.6726 & 0.7516 & 0.6768  \\
PatchCore~\cite{roth2022towards}             & 0.7058 & 0.7434 & 0.4605  \\
STFPM~\cite{wang2103student}                 & 0.5657 & 0.7371 & 0.6292  \\
CFlow~\cite{gudovskiy2022cflow}                 & 0.5433 & 0.7025 & 0.5850   \\
GANomaly~\cite{akcay2019ganomaly}               & 0.6239 & 0.7823 & 0.6871  \\
GANomaly\_1d           & 0.5913 & 0.7166 & 0.6884  \\  
\textbf{Ours}                  & \textbf{0.8658} & \textbf{0.8458} & \textbf{0.7292}  
\end{tblr}}
\caption{Anomaly detection AUROC of state-of-the-art methods on DataCon2020, ISCX VPN and non-VPN, UNB-CIC Tor and non-Tor datasets. We set up different random seeds for three experiments to obtain the average results. Our method achieves the best detection performance on each dataset.
}
\label{table:compare with others}
\end{table}

\section{Experimental Results}
\label{others}

\subsection{Datasets}

We have selected three widely used network traffic datasets for our experiments.

The "UNB-CIC Tor and non-Tor" dataset, captured by Arash et al.~\cite{lashkari2017characterization}, is collected using Wireshark and Tcpdump. The dataset includes both regular and Tor traffic captured from the workstation and gateway, encompassing 14 categories of traffic such as Chat, Streaming, Email, and others.

The "ISCX VPN and non-VPN" dataset, captured by Gerard et al.~\cite{draper2016characterization}, is collected using Wireshark and tcpdump.
During the capturing process, only the packets with the target IP were captured. The dataset comprises a total of 14 categories for regular and VPN traffic, including File Transfer, P2P, and more.

\begin{table}
\centering
\begin{tblr}{
  cells = {c},
  cell{1}{1} = {r=2}{},
  vline{2} = {1-11}{},
  hline{1,3,11-12} = {-}{},
}
Model                 & Params & FLOPs \\
                      & (M)    & (G)   \\
PADIM~\cite{defard2021padim}                 & \textbf{2.78}   & \underline{0.05}  \\
GANomaly~\cite{akcay2019ganomaly}              & 10.73  & 0.65  \\
GANomaly\_1d          & 45.89  & 11.94 \\
FastFlow~\cite{yu2021fastflow}              & 7.46   & 0.13  \\
DRAEM~\cite{zavrtanik2021draem}                 & 97.43  & 3.11  \\
Reverse Distillation~\cite{deng2022anomaly} & 80.61  & 0.61  \\
CFlow~\cite{gudovskiy2022cflow}                 & 6.45   & 0.15  \\
STFPM~\cite{wang2103student}                 & 5.57   & 0.10  \\
Ours              & \underline{3.91}   & \textbf{0.02}  
\end{tblr}
\caption{Comparison of model parameters and FLOPs in the inference phase. Our approach has the lowest FLOPs and the best detection performance, while also having small model parameter sizes. The effectiveness of the PADIM method depends on the performance of the pre-trained model employed. The size of the model will increase as the capability of the pre-trained model increases.}
\label{model size}
\end{table}

The "DataCon2020" dataset~\cite{DataCon2020} is derived from malicious and benign software collected between February and June 2020.
The traffic was generated by sandbox collection from Qi An Xin. 
The dataset defines malicious traffic as encrypted traffic generated by malware, while the benign traffic represents encrypted traffic generated by benign software.

Our datasets include encrypted and unencrypted traffic datasets, benign traffic and malicious traffic datasets.
In some scenarios, encrypted traffic is not allowed to be used, so naturally we define encrypted traffic and malicious traffic as anomaly traffic, and non-encrypted traffic and benign traffic as normal traffic.
Our training set consists only of normal samples. 
We have randomly selected 10,000 normal samples from one dataset to form our training set. Subsequently, we have obtained our testing set by randomly selecting 5,000 samples from the remaining pool of normal samples and another 5,000 samples randomly drawn from the abnormal samples. This process is also replicated for the training and testing sets of the other two datasets.

\subsection{Pre-processing}
\label{section:pre-processing}

\paragraph{Traffic cleaning}
We first remove packets in the datasets that may introduce confusion in anomaly detection.
The DNS protocol is used to map domain names and IP addresses to each other.
Both normal and anomaly traffic obtain corresponding IP addresses through the DNS protocol prior to conducting activities.
These traffic packets are not directly associated with normal and anomaly characteristics and do not contribute to anomaly detection.
The TCP protocol has a series of stages related to connections, including connection establishment, termination, and confirmation.
These packets often do not contain any actual payload, they are only associated with the connection and have no direct relevance to specific activities.
So we remove both types of packets~\cite{lotfollahi2020deep}.
In addition, there is also the ARP protocol, which is responsible for the mapping between IP addresses and MAC addresses, but it is often not directly associated with the activities of upper-layer users.
Therefore, we also remove packets related to the ARP protocol.

Then we process the structure information in the packet.
The packet header of the data link layer is often responsible for managing the transmission of specific physical links and cannot provide sufficient  information for anomaly detection.
Therefore, we remove the data link layer header.
In the header of the IP protocol, there are two fields, the destination IP address and the source IP address.
These two fields can summarize a series of data communications between hosts.
Detecting anomaly traffic only based on IP addresses is considered a shortcut rather than true learning. Therefore, we anonymize the IP addresses of all packets.

\paragraph{Traffic encoding}

We use byte encoding to process the packets by converting the individual bytes in the packets into corresponding decimal numbers. This allows us to obtain a one-dimensional array of packets.
To ensure uniformity and facilitate model comprehension, we fill the optional fields of the IP and TCP protocols.
In addition, there are two protocols, TCP and UDP, in the transport layer.
We have also filled the header of the UDP protocol to match the header length of TCP, ensuring consistency between the two protocols.
The neural network model requires us to unify the length of all packets.
Taking into account our statistical analysis of packet lengths and the preprocessing requirements for comparative experiments, we have determined the length of the one-dimensional packet array to be 1600 bytes.
Subsequently, we normalize the packets to a range of 0 to 1.
Finally, we combine the processed packets with the corresponding labels and store them in a CSV file format.

\subsection{Implementaion Details}

Our model is trained for 100 epochs with \emph{early stopping} techniques.
For the generator $G$ and discriminator $D$ in the feature extractor, we use two \emph{Adam} optimizers with a learning rate of 0.001, betas of 0.5 and 0.999 to train respectively.
The $w_{adv}$ in the feature extractor loss is set to 1, while $w_{rec}$ is set to 50. Additionally, we have determined the size of the hidden vector extracted by the feature extractor to be 70.
We utilize the bidirectional flow module, which consists of 8 coupling blocks. For training, we employ the \emph{Adam} optimizer with a learning rate of 0.001 and betas of 0.5 and 0.999.
The two parameters, $\mu$ and $\sigma$, used for generating noise are determined through experimental analysis.
When training the classifier, we utilize the \emph{Adam} optimizer with a learning rate of 0.001 and betas of 0.5 and 0.999. These parameters have been determined through an process of grid search and experiments.

\begin{table*}[ht]
\centering
\scalebox{0.9}{
\begin{tblr}{
  cells = {c},
  cell{1}{1} = {c=2}{},
  cell{1}{3} = {c=3}{},
  cell{1}{6} = {c=2}{},
  cell{1}{8} = {c=3}{},
  cell{3}{1} = {r=4}{},
  cell{3}{6} = {r=4}{},
  cell{7}{1} = {r=4}{},
  cell{7}{6} = {r=4}{},
  cell{11}{1} = {r=4}{},
  cell{11}{6} = {r=4}{},
  cell{15}{1} = {r=4}{},
  cell{15}{6} = {r=4}{},
  cell{19}{1} = {r=4}{},
  cell{19}{6} = {r=4}{},
  vline{2,7} = {2-22}{},
  vline{1-3,6,8,11} = {1-22}{},
  vline{3,6,8,11} = {4-6,8-10,12-14,16-18,20-22}{},
  hline{1-3,7,11,15,19,23} = {-}{},
}
Noise Distribution &     & Dataset         &                 &                 & Noise Distribution &     & Dataset &        &         \\
$\mu$                  & $\sigma$   & TOR             & VPN             & DataCon         & $\mu$                  & $\sigma$   & TOR     & VPN    & DataCon \\
-100               & 0.1 & 0.6344          & 0.4598          & 0.6850          & 5                  & 0.1 & 0.7695  & 0.7910 & 0.7058  \\
                   & 0.5 & 0.5888          & 0.5771          & 0.6906          &                    & 0.5 & 0.5749  & 0.8082 & 0.6985  \\
                   & 5   & 0.6766          & 0.7582          & \textbf{0.7292} &                    & 5   & 0.7004  & 0.7707 & 0.6828  \\
                   & 15  & 0.6051          & 0.7221          & 0.7137          &                    & 15  & 0.7418  & 0.8076 & 0.6926  \\
-25                & 0.1 & 0.6482          & 0.7609          & 0.4488          & 9                  & 0.1 & 0.6364  & 0.6703 & 0.6740  \\
                   & 0.5 & 0.7189          & 0.8191          & 0.5752          &                    & 0.5 & 0.6331  & 0.8160 & 0.7020  \\
                   & 5   & \textbf{0.8458} & 0.7732          & 0.7146          &                    & 5   & 0.6545  & 0.7989 & 0.6924  \\
                   & 15  & 0.8161          & 0.7891          & 0.7092          &                    & 15  & 0.6471  & 0.7427 & 0.6697  \\
-10                & 0.1 & 0.6553          & 0.8201          & 0.6893          & 10                 & 0.1 & 0.4361  & 0.8370 & 0.7207  \\
                   & 0.5 & 0.6138          & 0.8414          & 0.6023          &                    & 0.5 & 0.6484  & 0.4467 & 0.7026  \\
                   & 5   & 0.5583          & 0.8602 & 0.6494          &                    & 5   & 0.7855  & 0.7529 & 0.6966  \\
                   & 15  & 0.7696          & 0.7892          & 0.5721          &                    & 15  & 0.6094  & 0.8002 & 0.6836  \\
-9                 & 0.1 & 0.6526          & 0.8118          & 0.6971          & 25                 & 0.1 & 0.5958  & 0.6327 & 0.6486  \\
                   & 0.5 & 0.7909          & 0.8155          & 0.6710          &                    & 0.5 & 0.4326  & 0.7602 & 0.7101  \\
                   & 5   & 0.5763          & \textbf{0.8658}          & 0.3903          &                    & 5   & 0.8331  & 0.5499 & 0.6550  \\
                   & 15  & 0.6485          & 0.7780          & 0.7062          &                    & 15  & 0.5875  & 0.8542 & 0.6755  \\
-5                 & 0.1 & 0.6523          & 0.7979          & 0.6531          & 100                & 0.1 & 0.7961  & 0.1944 & 0.6966  \\
                   & 0.5 & 0.5938          & 0.7042          & 0.5486          &                    & 0.5 & 0.6590  & 0.7584 & 0.6836  \\
                   & 5   & 0.6146          & 0.4137          & 0.6834          &                    & 5   & 0.5877  & 0.8143 & 0.6998  \\
                   & 15  & 0.7199          & 0.7368          & 0.6655          &                    & 15  & 0.5606  & 0.8347 & 0.6805  
\end{tblr}
}
\caption{Detection performance on different noise distributions. The noise distribution determines the generated pseudo anomaly samples. By the trick of reparameterization, we can easily get different noise distributions. We explore the effects of different noise distributions on the three datasets.}
\label{table:different noise distribution}
\end{table*}

\subsection{Comparison with Existing Methods}

We analyze anomaly network traffic through our framework of anomaly detection.
To demonstrate the superiority of our framework, we extensively have compared it with state-of-the-art methods in the field of anomaly detection, including knowledge distillation based~\cite{deng2022anomaly,wang2103student}, reconstruction based~\cite{akcay2019ganomaly,zavrtanik2021draem}, normalization flow based~\cite{yu2021fastflow,gudovskiy2022cflow}, memory matching based~\cite{roth2022towards}, and distribution learning based methods~\cite{ahuja2019probabilistic,zavrtanik2021draem,defard2021padim}.
The methods used for comparison can be found in~\cite{anomalib}, including the code and settings employed.
For the sake of fairness, We evaluate some methods on the three datasets by transforming the preprocessed packets from a one-dimensional structure to a two-dimensional grayscale image.
Since these methods employ pre-trained feature extractors specifically designed for two-dimensional images.
In addition, we have also modified the model in~\cite{akcay2019ganomaly} and obtained a new method specifically for processing one-dimensional structured packets, which we refer to as GANomaly\_1d provided within the code.
We run experiments 5 times with different random seeds and report the mean AUC.

As shown in Table \ref{table:compare with others}, we can clearly see that on the "UNB-CIC Tor and non-Tor" and "ISCX VPN and non-VPN" datasets, our method achieves the best results and leads the second-ranked methods with a significant advantage of 6\% and 15\% respectively.
In the knowledge distillation based anomaly detection method~\cite{deng2022anomaly,wang2103student}, the difference between the representations of the anomaly samples by the teacher model and the student model is used to detect the anomaly.  However, the detection is not effective because the pre-trained teacher model on ImageNet~\cite{deng2009imagenet} has not seen the network traffic samples.
Additionally, the conversion of the packets from their original one-dimensional semantic structure to a two-dimensional format disrupts their inherent structure. The added dimension does not add any additional information.
Feature extraction is commonly employed in anomaly detection~\cite{pang2021deep}.
Our method employs the feature extractor that is pre-trained specifically on normal one-dimensional network traffic packets.
We design the unsupervised feature extractor to represent network traffic packets effectively. This approach avoids directly generating raw packets, which would destroy the data structure and produce meaningless samples, as demonstrated in the Section~\ref{ablating feature extractor}.
The extracted features are more beneficial for enhancing the performance of downstream anomaly detection tasks.

\begin{figure*}[ht]
  \centering
  \subfloat[Tor and non-Tor dataset]{
    \includegraphics[width=0.3\textwidth]{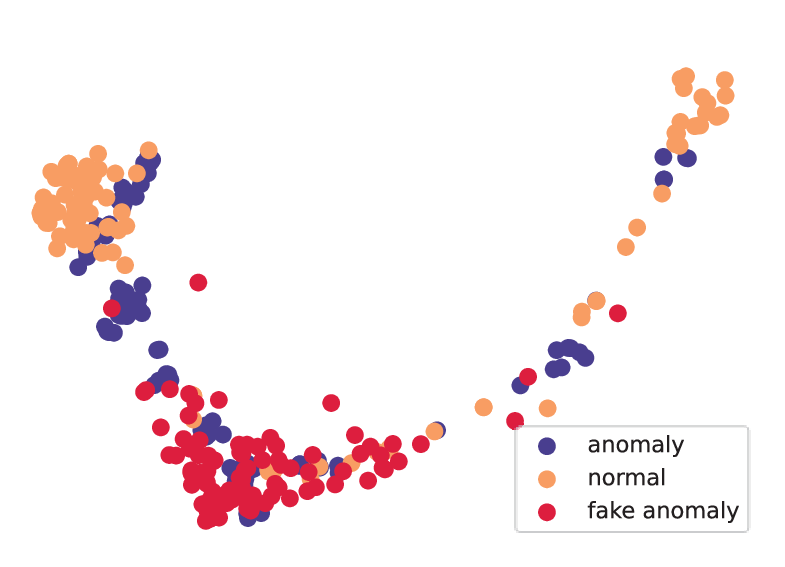}
  }
  \subfloat[VPN and non-VPN dataset]{
    \includegraphics[width=0.3\textwidth]{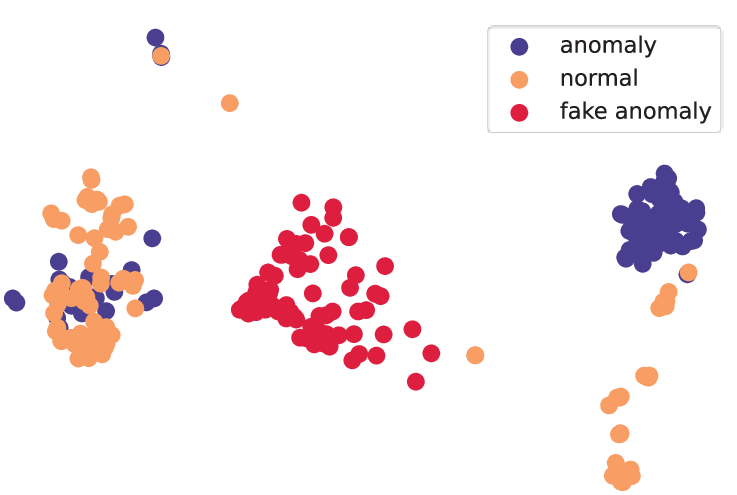}
  }
  \subfloat[DataCon2020 dataset]{
    \includegraphics[width=0.3\textwidth]{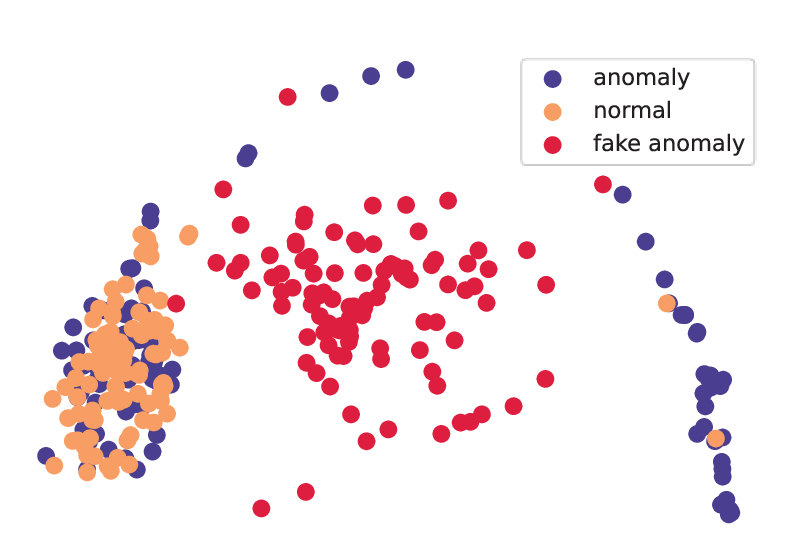}
    }
    \caption{T-SNE visualization of representations in latent space. We plot the features of the normal, anomaly, and synthetic anomaly samples. It can be seen that our synthetic anomaly samples do not overlap well with real anomaly samples, but they are significantly different from normal samples. The model learns how to accurately identify normal traffic by distinguishing between them.
    }
    \label{fig:t-sne}
\end{figure*}

A traffic packet is a one-dimensional data structure containing specific semantic fields but lacks spatial semantics~\cite{zheng2022multi}.
For methods that employ generated anomaly samples for anomaly detection, such as DREAM~\cite{zavrtanik2021draem}, introducing noise into normal packet images is meaningless as it destroys the semantic fields of the original packet and does not effectively generate anomaly samples, resulting in poor results.
Some of the problems that exist in current network traffic generation research~\cite{xu2021stan,yin2022practical}, such as the difficulty of training GANs and the need to collect a large number of samples.
Our method is not affected by these issues.
Our method maps the normal samples to the space of standard normal distribution, and simulates the anomaly samples through manipulating them in this space.
This method avoids disrupting the specific semantic fields of the original data packet and allows for a closer semantic alignment with real anomaly samples.
At the same time, we do not need to collect anomaly samples or a priori knowledge of anomaly patterns, but only simulate anomaly samples by normal samples and randomly sampled noise.
By adjusting the distribution of random noise, we are able to simulate different anomaly samples.
On the "DataCon2020" dataset, our method also achieves the best results, but the overall performance of all methods is not particularly outstanding.
We analyze that the diversity of categories and the insufficient number of samples in each category in this dataset have proposed challenges for each detection method.
In addition, as shown in Figure~\ref{fig:t-sne}, the similarity between normal packet samples and abnormal packet samples in the DataCon dataset is relatively high, posing a challenge for the model to detect abnormal samples.

As shown in Figure~\ref{fig:anomaly_score}, we draw the histogram of the detection results on the test sets of the three datasets to visualize the detection effect of our model.
Our model demonstrates excellent performance in distinguishing between normal and anomaly traffic on both the "UNB-CIC Tor
and non-Tor" and the "ISCX VPN and non-VPN" datasets.
Additionally, on the "DataCon2020 dataset", our model currently exhibits the best detection performance, although there is still potential for further improvement.

We also compare the model sizes and the FLOPs of the different methods during the inference, as shown in Table~\ref{model size}.
Our approach is effective for detecting network traffic anomalies in computer power limited deployment environments.
After training our model, anomaly detection can be achieved by retaining only the encoder part of the feature extractor and the classifier in the inference process, and the rest of the modules are only used to support the training.
Unlike other normalizing flow based methods~\cite{yu2021fastflow,gudovskiy2022cflow}, our approach does not require the flow module to compute anomaly scores, and serves as a module for synthesizing anomaly samples during the training process.

We employ t-SNE to project the features of the normal, anomaly, and synthetic anomaly samples into a 2D space, as showed in Figure~\ref{fig:t-sne}.
Ideally, we hope that the simulated anomaly samples overlap well with the real anomaly samples. However, this requires us to spend more time carefully designing the distribution of the noise. 
For simplicity, we have tried sampling the noise from a random Gaussian distribution and changing the normal sample properties with it.
While the synthetic anomaly features may not perfectly simulate real anomaly samples, the model improves the discrimination of normal samples by distinguishing between normal and synthetic anomaly samples.

\begin{table*}[htb]
\centering
\scalebox{0.8}{
\begin{tblr}{
  cells = {c},
  cell{1}{2} = {c=2}{},
  cell{1}{4} = {c=3}{},
  cell{1}{8} = {c=2}{},
  cell{1}{10} = {c=3}{},
  cell{3}{1} = {r=15}{},
  cell{3}{7} = {r=15}{},
  vline{1-2,4,7-8,10,13} = {1-18}{},
  vline{1-4,7-10,13} = {2-3,18}{},
  vline{3-4,7,9-10,13} = {4-17}{},
  hline{1,19} = {-}{0.08em},
  hline{2-3,18} = {-}{},
}
                                       & Noise Distribution &       & Dataset         &                 &                 &                                                     & Noise Distribution &       & Dataset         &                 &                 \\
                                       & $\mu$                  & $\sigma$     & VPN             & TOR             & DataCon         &                                                     & $\mu$                  & $\sigma$     & VPN             & TOR             & DataCon         \\
\begin{sideways}w/o Bidirectional Flow Module\end{sideways} & -100               & 5     & 0.7557          & 0.5573          & 0.6768          & \begin{sideways}w/o Feature Extractor\end{sideways} & -100               & 5     & 0.7325          & 0.5711          & 0.7169          \\
                                       & -25                & 5     & 0.5324          & 0.6042          & 0.6705          &                                                     & -25                & 5     & 0.5696          & 0.4260          & 0.3805          \\
                                       & -20                & 10    & 0.7736          & 0.4758          & 0.3844          &                                                     & -20                & 10    & 0.6315          & 0.5567          & 0.6728          \\
                                       & -10                & 5     & 0.3972          & 0.5509          & 0.6723          &                                                     & -10                & 5     & 0.3749          & 0.4729          & 0.6181          \\
                                       & -10                & 1     & 0.3746          & 0.4381          & 0.3440          &                                                     & -10                & 1     & 0.6870          & 0.5819          & 0.5431          \\
                                       & -9                 & 5     & 0.6804          & 0.6120          & 0.3334          &                                                     & -9                 & 5     & 0.6680          & 0.4549          & 0.6620          \\
                                       & -9                 & 1     & 0.4185          & 0.6638          & 0.4544          &                                                     & -9                 & 1     & 0.6383          & 0.4909          & 0.5789          \\
                                       & -5                 & 5     & 0.5244          & 0.7459          & 0.4250          &                                                     & -5                 & 5     & 0.6515          & 0.4481          & 0.6768          \\
                                       & 5                  & 1.5   & 0.3090          & 0.5880          & 0.6427          &                                                     & 5                  & 1.5   & 0.6810          & 0.5010          & 0.6783          \\
                                       & 5                  & 15    & 0.5873          & 0.6091          & 0.6594          &                                                     & 5                  & 15    & 0.7976          & 0.4838          & 0.4110          \\
                                       & 9                  & 0.1   & 0.7900          & 0.6344          & 0.6705          &                                                     & 9                  & 0.1   & 0.2565          & 0.5016          & 0.6624          \\
                                       & 9                  & 1     & 0.7767          & 0.5591          & 0.6760          &                                                     & 9                  & 1     & 0.3284          & 0.5169          & 0.6785          \\
                                       & 10                 & 5     & 0.4010          & 0.6177          & 0.3828          &                                                     & 10                 & 5     & 0.3127          & 0.4883          & 0.6603          \\
                                       & 20                 & 1     & 0.2760          & 0.5943          & 0.6770          &                                                     & 20                 & 1     & 0.5016          & 0.4624          & 0.6800          \\
                                       & 25                 & 5     & 0.6552          & 0.4819          & 0.6297          &                                                     & 25                 & 5     & 0.5243          & 0.6912          & 0.6745          \\
Ours                                   & -9/-25/-100        & 5/5/5 & \textbf{0.8658} & \textbf{0.8458} & \textbf{0.7292} & Ours                                                & -9/-25/-100        & 5/5/5 & \textbf{0.8658} & \textbf{0.8458} & \textbf{0.7292}           
\end{tblr}
}
\caption{Detection performance on ablating different modules of our method. We conduct ablation experiments on the bidirectional flow module and the feature extractor to demonstrate the effectiveness of our proposed modules. The bottom row corresponds to the results of our method with both modules, where $\mu$ and $\sigma$ represent different datasets, respectively.}
\label{table:ablate module}
\end{table*}

\begin{table*}[ht]
\centering
\scalebox{0.8}{
\begin{tblr}{
  cells = {c},
  cell{1}{2} = {c=2}{},
  cell{1}{4} = {c=3}{},
  cell{1}{8} = {c=2}{},
  cell{1}{10} = {c=3}{},
  cell{3}{1} = {r=15}{},
  cell{3}{7} = {r=15}{},
  vline{1-2,4,7-8,10,13} = {1-18}{},
  vline{1-3,5,8-9,11} = {1}{},
  vline{1-4,7-10,13} = {2-3,18}{},
  vline{3-4,7,9-10,13} = {4-17}{},
  hline{1-3,18-19} = {-}{},
}
                                                 & Noise Distribution &       & Dataset         &                 &                 &                                                  & Noise Distribution &       & Dataset         &                 &                 \\
                                                 & $\mu$                  & $\sigma$     & VPN             & TOR             & DataCon         &                                                  & $\mu$                  & $\sigma$     & VPN             & TOR             & DataCon         \\
\begin{sideways}anomaly:normal=1:1\end{sideways} & -100               & 5     & 0.3333          & 0.6416          & 0.6771          & \begin{sideways}anomaly:normal=2:1\end{sideways} & -100               & 5     & 0.2876          & 0.3372          & 0.6693          \\
                                                 & -25                & 5     & 0.8419          & 0.5984          & 0.6985          &                                                  & -25                & 5     & 0.6676          & 0.6493          & 0.6827          \\
                                                 & -20                & 10    & 0.7559          & 0.6317          & 0.6945          &                                                  & -20                & 10    & 0.7154          & 0.4850          & 0.7059          \\
                                                 & -10                & 5     & 0.7898          & 0.6274          & 0.6837          &                                                  & -10                & 5     & 0.5435          & 0.5279          & 0.6670          \\
                                                 & -10                & 1     & 0.6979          & 0.6394          & 0.7031          &                                                  & -10                & 1     & 0.3853          & 0.7602          & 0.6894          \\
                                                 & -9                 & 5     & 0.8019          & 0.6781          & 0.6895          &                                                  & -9                 & 5     & 0.8445          & 0.5985          & 0.6731          \\
                                                 & -9                 & 1     & 0.7897          & 0.7820          & 0.6981          &                                                  & -9                 & 1     & 0.8083          & 0.6944          & 0.6915          \\
                                                 & -5                 & 5     & 0.7233          & 0.4607          & 0.6775          &                                                  & -5                 & 5     & 0.7969          & 0.6685          & 0.6545          \\
                                                 & 5                  & 1.5   & 0.7899          & 0.7173          & 0.7137          &                                                  & 5                  & 1.5   & 0.8115          & 0.5829          & 0.6892          \\
                                                 & 5                  & 15    & 0.5487          & 0.5469          & 0.6912          &                                                  & 5                  & 15    & 0.6553          & 0.5507          & 0.5893          \\
                                                 & 9                  & 0.1   & 0.7639          & 0.5671          & 0.7001          &                                                  & 9                  & 0.1   & 0.7880          & 0.6879          & 0.6967          \\
                                                 & 9                  & 1     & 0.3962          & 0.7811          & 0.6576          &                                                  & 9                  & 1     & 0.8460          & 0.6487          & 0.6999          \\
                                                 & 10                 & 5     & 0.8308          & 0.6906          & 0.6855          &                                                  & 10                 & 5     & 0.7889          & 0.5456          & 0.6831          \\
                                                 & 20                 & 1     & 0.7555          & 0.7493          & 0.6429          &                                                  & 20                 & 1     & 0.6377          & 0.7505          & 0.7047          \\
                                                 & 25                 & 5     & 0.8197          & 0.6163          & 0.5783          &                                                  & 25                 & 5     & 0.7645          & 0.6256          & 0.6793          \\
Ours                                             & -9/-25/-100        & 5/5/5 & \textbf{0.8658} & \textbf{0.8458} & \textbf{0.7292} & Ours                                             & -9/-25/-100        & 5/5/5 & \textbf{0.8658} & \textbf{0.8458} & \textbf{0.7292} 
\end{tblr}
}

\caption{Detection performance on different ratios of normal samples to abnormal samples. We achieve different ratios by altering the quantity of pseudo anomaly samples during the training process. The bottom row shows our method, where pseudo abnormal samples account for half of normal samples, where $\mu$ and $\sigma$ represent different datasets, respectively.}
\label{table:different sample}
\end{table*}

\subsection{Ablation Study}
\label{ablation study}

\subsubsection{Adopting Different Noise Distributions}
For our method, the noise distribution plays a particularly critical role, as it directly determines the quality of the generated samples.
In~\cite{kingma2018glow}, the authors determine the direction of attribute change by the difference between two sample vectors with specific attributes. And since our model is only trained on normal samples, we can only guide the generation of simulated abnormal samples by trying random noise. This aspect is crucial for training classifiers effectively.

We have tried different combinations of $\mu$ and $\sigma$, and the experimental results are shown in Table~\ref{table:different noise distribution}. 
This shows that our model is sensitive to the parameters of the noise distribution, which is in line with our expectations.
The distribution of the simulated anomaly samples in the standard normal distribution space is difficult to determine, depending on the noise distribution from which the noise is sampled.
We simulate the anomaly samples by manipulating the normal samples, thereby enabling the classifier to show excellent recognition capability for normal samples.
In addition, we can easily generate various types of anomaly samples only by changing noise distributions.

\subsubsection{Ablating Bidirectional Flow Module}
Generating anomalous samples is crucial to our approach, and the Bidirectional Flow Module is able to fit more complex distributions by combining multiple invertible transformations, resulting in high quality synthetic samples. 
In addition, by training this module we are able to map normal samples to a specified distribution, while anomaly samples not seen by the module will be mapped outside of the distribution, thus simulating abnormal samples by deviating from the normal samples.

We remove the bidirectional flow module and directly introduce noise to the latent vectors extracted by the feature extractor to simulate anomaly samples. 
Then, the simulated samples are fed into the classifier for detection.
We explore different combinations of $\mu$ and $\sigma$, and the experimental results are shown in Table~\ref{table:ablate module}.
It can be observed that directly introducing noise to the latent vectors to simulate anomaly samples does not achieve satisfactory detection results.
The latent vectors can affect the properties of the samples, but we have difficulty in achieving the semantic conversion from normal to abnormal samples by random noise, which may require careful design of the vectors.
Manipulating vectors in the standard normal space proves to be more effective to altering the properties of network traffic.

\subsubsection{Ablating Feature Extractor}
\label{ablating feature extractor}
We conducted training using normal samples to acquire a feature extractor capable of effective feature extraction on packets of network traffic. 
Autoencoders are often used for extracting features from network traffic~\cite{aceto2019mobile,ring2019flow}.
However, these features are typically deep representations of manually extracted features. This approach may overlook potential data connections within the packets.
Our feature extractor introduces a discriminator to improve the capability of the autoencoder, which is more helpful in generating dense sample features.

We remove the feature extractor in our feature extractor ablation experiments and directly feed the preprocessed one-dimensional packets into the bidirectional flow module.
The bidirectional flow module is trained to map the one-dimensional packet vectors to the standard normal distribution space. 
In the standard normal distribution space, noise is introduced to the vectors, and through the generation direction, simulated anomaly samples are obtained. 
Both normal samples and simulated anomaly samples are then fed into the classifier for detection.
The results obtained by retraining the parameters are shown in Table~\ref{table:ablate module}.
The detection performance directly using one-dimensional packet vectors is poor on these three datasets.
We believe that the packet vectors without feature extraction may contain a significant amount of redundant information, which hampers model learning by lacking concentrated information. Consequently, this leads to poor performance.

\subsubsection{Adjusting the Ratio of Samples}
For our model, the aim is for the classifier to learn how to distinguish normal samples with a learning focus on such samples. In addition, we think that if the number of synthetic abnormal samples is increased, it is possible that the model will shift its learning focus.

We adjust the number of generated anomaly samples to demonstrate the impact of different ratios of normal samples to anomaly samples on the model.
In our method, the number of synthetic anomaly samples is half the number of normal samples.
The other two comparative methods involve maintaining an equal number of normal and anomaly samples, and having twice the number of anomaly samples compared to normal samples.
The experimental results are shown in Table \ref{table:different sample}.
Compared to our method, when the number of anomaly samples increases to be equal to the number of normal samples, there is a slight decrease in performance on all three datasets.
As the number of anomaly samples continues to increase to twice the number of normal samples, the performance still decreases.
Changing the ratio of samples will have an impact on the classifier. 
We expect the classifier to distinguish between normal and anomaly samples. 
However, when the number of anomaly samples increases, the classifier tends to pay excessive attention to the anomaly samples, resulting in a decrease in detection performance.


\begin{table}[htb]
\centering
\scalebox{1}{
\begin{tblr}{
  cells = {c},
  cell{1}{1} = {c=2}{},
  cell{1}{3} = {c=2}{},
  cell{1}{5} = {r=2}{},
  vline{2,4} = {1}{},
  vline{2-5} = {2-6}{},
  vline{3,5} = {1}{},
  hline{1,3,5,7} = {-}{},
  hline{2} = {1-4}{},
}
Train ~ &     & Test ~  &     & Result ~ \\
DataCon & TOR & DataCon & TOR &          \\
$\surd$       & ~   & $\surd$       & ~   & 0.7292   \\
$\surd$       & ~   & ~       & $\surd$   & 0.7058   \\
~       & $\surd$   & ~       & $\surd$   & 0.8458   \\
~       & $\surd$   & $\surd$       & ~   & 0.8060   
\end{tblr}
}
\caption{Detection performance across the datasets. "UNB-CIC Tor and non-Tor" is the encrypted and unencrypted traffic dataset, and the "DataCon2020" dataset is the benign and malicious traffic dataset. Comparing with training and testing on the same dataset, the detection performance across the datasets have a slightly decrease.}
\label{table:generalization}
\end{table}

\subsubsection{Generalizing across the Datasets}
In a real network traffic detection scenario, the trained model will be faced with a large number of unknown network traffic packets, both normal and anomaly traffic. We will explore the detection ability of the model when faced with the test samples from an unknown distribution, so we further investigate the generalization ability across datasets.

While the anomaly samples in three datasets  are different, the normal samples are all captured from normal activities and have similarities. We further study the generalization of our method across different datasets.
Both the "UNB-CIC Tor and non-Tor" and "ISCX VPN and non-VPN" datasets consist of encrypted and unencrypted traffic.
We extend our experiments on one of these datasets and "DataCon2020" dataset.
We train our model on one dataset and test on another dataset to assess its ability to generalize to unseen anomaly samples.
As shown in Table~\ref{table:generalization}, when training our model on one dataset and testing it on another, the results show a slight drop, but remain within acceptable limits.
This shows that our method can have relatively good detection results even in the presence of unknown anomaly traffic in real detection environments.
Our approach improves the model's ability to identify normal network traffic by classifying pseudo anomaly traffic. It is effective across different data distributions.

\section{Conclusion}
In this paper, we propose a three-stage framework for anomaly traffic detection that involves generating simulated anomaly samples.
Our approach is able to generate anomaly samples with unknown patterns, without prior knowledge of the anomalies, and use them to improve anomaly detection.
The key lies in the feature extractor and bidirectional flow module designed specifically for traffic. 
These modules enable us to transform the packets into the standard normal distribution space, where we manipulate the vectors to alter the properties of the traffic packets.
This technique allows us to simulate the anomaly traffic.
Our method demonstrates excellent performance in anomaly detection across three real network traffic datasets.
We envision that our method of constructing anomaly samples can be widely applied in many fields, serving as a reliable technique for generating simulated anomaly samples.

\bibliographystyle{ieeetr}
\bibliography{Example.bbl}

\vfill
\end{document}